%% file: camera_ready_v3_shuwei.tex
\documentclass[10pt,twocolumn,letterpaper]{article}

\usepackage{cvpr}      

\input{preamble}

\definecolor{cvprblue}{rgb}{0.21,0.49,0.74}
\usepackage[pagebackref,breaklinks,colorlinks,allcolors=cvprblue]{hyperref}

\def\paperID{} 
\def\confName{CVPR}
\def\confYear{2026}

\title{White-Balance First, Adjust Later: Cross-Camera Color Constancy via Vision-Language Evaluation}

\author{Shuwei Li$^1$~~~~~~~~~Lei Tan$^1$~~~~~~~~~Robby T. Tan$^{1,2}$\thanks{Robby T. Tan was affiliated with ASUS Intelligent Cloud Services during part of the period in which this work was developed.}\\
$^1$National University of Singapore~~~~~~~$^2$ASUS Intelligent Cloud Services\\
{\tt\small shuwei@u.nus.edu~~let.tan@nus.edu.sg~~robby.tan@nus.edu.sg}
}

\begin{document}

\twocolumn[{%
  \maketitle
  \renewcommand\twocolumn[1][]{#1}%
  \begin{center}
    \includegraphics[width=\textwidth]{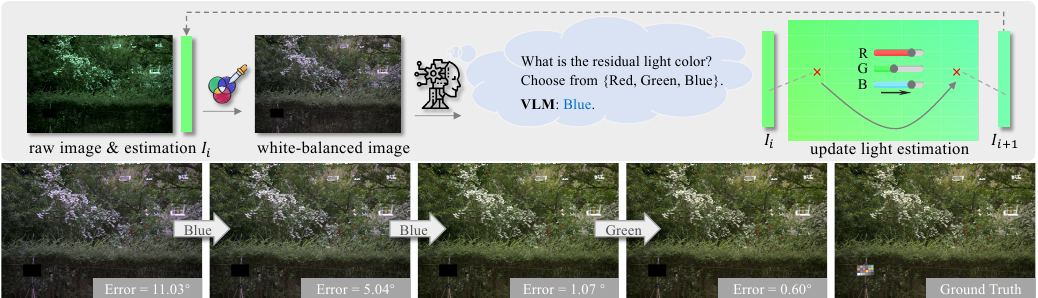}
    \captionof{figure}{Rather than directly predicting a light color, our method first white-balances the image then later updates the light estimate via semantic feedback. Given a raw image, we white-balance the image with the current estimate and ask a vision-language model to identify the residual color cast. The predicted cast induces a directional update in chromacity space, yielding a refined estimate for the next iteration. Bottom: an example sequence with angular error decreasing from \(11.03^\circ\) to \(0.60^\circ\); the rightmost image shows the ground truth.}
    \label{fig:head}
  \end{center}%
}]

\begingroup
\renewcommand\thefootnote{}
\footnotetext{*Robby T. Tan was affiliated with ASUS Intelligent Cloud Services during part of the period in which this work was developed.}
\endgroup

\begin{abstract}
Color constancy aims to keep object colors consistent under varying illumination. 
Cross-camera generalization in color constancy remains challenging because learning-based models often overfit to the color response characteristics of the training camera, resulting in degraded performance on images captured by other cameras.
We propose \textbf{VLM-CC}, a feedback-guided framework that formulates color constancy as an iterative refinement process.
Instead of directly estimating the illuminant from raw input, VLM-CC performs iterative correction driven by vision-language model (VLM)-based evaluation.
At each iteration, the image is white-balanced using the current estimate and converted to pseudo-sRGB. 
A lightweight LoRA-tuned VLM then assesses the corrected image, identifying the dominant residual color cast and providing qualitative feedback.
This feedback is mapped to a residual illumination direction (red, green, or blue) and used to update the illuminant estimate until convergence.
Our key idea is to reframe color constancy as an iterative perceptual feedback problem, leveraging VLM evaluation instead of direct RGB regression.
By replacing direct RGB estimation with VLM-guided perceptual feedback, VLM-CC achieves state-of-the-art robustness in cross-camera color constancy across multiple datasets.
Code will be available at \url{https://github.com/NothingIknow/VLM-CC}.
\end{abstract}

\section{Introduction}
\label{sec:intro}
Color constancy aims to maintain consistent object colors under varying illumination, ensuring that perceived colors remain stable regardless of lighting changes.
%
Over the past decades, many computational color constancy (CC) methods have been developed, which can broadly be categorized into physics-based, statistical, and learning-based methods.

Physics-based methods (e.g.,~\cite{tominaga1996multichannel,finlayson2001convex,finlayson2001solving,woo2017improving}) infer illumination by modeling the image formation process and applying constraints on reflectance, shading, and illumination.
They are interpretable and data-independent but rely on simplified assumptions about scene geometry and lighting.
Statistical methods (e.g.,~\cite{van2007edge,qian2019cvpr,land1971lightness,chakrabarti2011color,brainard1986analysis}) instead exploit global or local color statistics, such as the gray world~\cite{buchsbaum1980spatial}, white patch~\cite{brainard1986analysis}, or gray edge~\cite{van2007edge}, but their accuracy significantly depends on scene color diversity.

Learning-based methods (e.g.,~\cite{lo2021clcc,afifi2022auto,bianco2019quasi,sidorov2019conditional,gehler2008bayesian, gijsenij2010color, yu2020cascading, barron2015convolutional, barron2017fast, hu2017fc4, li2024nightcc, tang2022transfer, afifi2025optimizing}) have recently achieved superior performance over physics- and statistics-based approaches by leveraging large collections of labeled image-illuminant pairs.
These models learn to regress illumination directly from raw data, capturing complex nonlinear relationships.
However, most learning-based methods are developed under a single-camera setting, where the image formation pipeline of the training camera remains fixed.
Such assumptions often lead to camera-specific overfitting, causing performance degradation when applied to different cameras.

To address cross-camera generalization, few-shot methods~\cite{mcdonagh2018formulating, hernandez2020multi} use a small labeled set of target-camera images with ground-truth illuminants, while DMCC~\cite{yue2024effective} instead relies on a single D65 white-point measurement per sensor. 
Other approaches such as SIIE~\cite{afifi2019sensor} learn a device-independent working space for raw images, while C5~\cite{afifi2021cross} utilizes unlabeled target-camera raw images. 
More recent methods like GCC~\cite{chang2025gcc} and CCMNet~\cite{kim2025ccmnet} remove the need for target-camera data at test time by exploiting generative diffusion priors or pre-calibrated parameters.

In contrast, human observers rarely rely on raw pixels alone when judging white-balance performance~\cite{hansen2006memory,olkkonen2008color,witzel2011object}.
When adjusting white balance manually, humans often refer to recognizable objects such as paper, skin, or sky, and iteratively correct the scene until it looks neutral.
This suggests that reliable color constancy benefits from semantic understanding of the scene. Moreover, evaluating the white-balanced result, rather than predicting from raw pixels in a single pass, is key to correcting inaccurate estimates.

In this paper, we propose \textbf{VLM-CC}, a feedback-based framework that leverages vision-language models (VLMs) for cross-camera color constancy.
As shown in Fig.~\ref{fig:head}, instead of regressing the illuminant directly from raw images, VLM-CC reformulates the problem as an iterative process guided by semantic feedback.
The process starts with an initial estimate from any simple method (e.g., Gray-World~\cite{buchsbaum1980spatial}).
Then, at each iteration, the raw image is white-balanced using the current estimate and converted into a pseudo-sRGB space, an approximation of true sRGB rendering since the illuminant is not yet fully corrected.
A LoRA-tuned VLM then analyzes the pseudo-sRGB image, identifying scene content and intrinsic object colors to derive a coarse color prior.
Conditioned on this prior, the VLM predicts a dominant residual color cast label from {red, green, blue}, which is mapped to a directional update of the illuminant estimate.
Through this iterative feedback, VLM-CC progressively refines the illumination and achieves robust cross-camera generalization without requiring camera calibration or retraining.
Our key novelty lies in recasting color constancy as a reasoning-driven feedback process, where a vision-language model acts as a perceptual evaluator that provides qualitative guidance rather than direct pixel-level regression, enabling better sensor invariance and interpretability.

In summary, our main contributions are as follows:
\begin{itemize}
\item We propose VLM-CC, a feedback-based color constancy framework that leverages a vision-language model to provide semantic feedback, evaluating whether the current white-balanced image still exhibits a red, green, or blue cast, to guide iterative correction across different cameras.
\item We reformulate color constancy as an iterative feedback process in pseudo-sRGB space: the raw image is white-balanced using the current estimate, converted to pseudo-sRGB via the camera's color matrix, and evaluated by a LoRA-tuned VLM that predicts the dominant residual cast (red, green, or blue) to update the illumination direction until convergence.
\item We demonstrate state-of-the-art cross-camera performance on NUS-8, Cube+, INTEL-TAU, and Gehler-Shi, with consistent improvements in mean angular error and robustness over recent strong baselines.
\end{itemize}

\section{Related Work}
\label{sec:related}
\noindent\textbf{Computational Color Constancy}
Computational color constancy focuses on ensuring that the perceived colors of objects remain consistent under varying lighting conditions~\cite{barnard1995computational}. In the context of digital photography, this procedure is commonly termed Auto White Balance (AWB). It is a core module in a camera’s ISP pipeline and in many raw-image workflows such as Photoshop~\cite{adobephotoshop}. AWB first estimates the scene light color, then applies a linear correction to the camera-specific raw image so that the result appears as if it were captured under neutral white light, and is closely related to broader low-level vision restoration tasks~\cite{jin2022unsupervised,jin2023enhancing,xiao2025incorporating,xiao2026learning}.

Statistical methods~\cite{land1977retinex,finlayson2004shades,van2007edge,gijsenij2011improving,cheng2014illuminant,qian2018revisiting,qian2019cvpr,land1971lightness,chakrabarti2011color,brainard1986analysis} estimate illumination based on assumptions about scene color distributions or spatial statistics. Representative examples include Gray-World~\cite{buchsbaum1980spatial}, White-Patch~\cite{brainard1986analysis}, and Shades-of-Gray~\cite{finlayson2004shades}. These approaches are simple and efficient, but their underlying assumptions often break in complex or unbalanced scenes, which limits their accuracy.

\begin{figure*}[t]
  \centering
  \includegraphics[width=\textwidth]{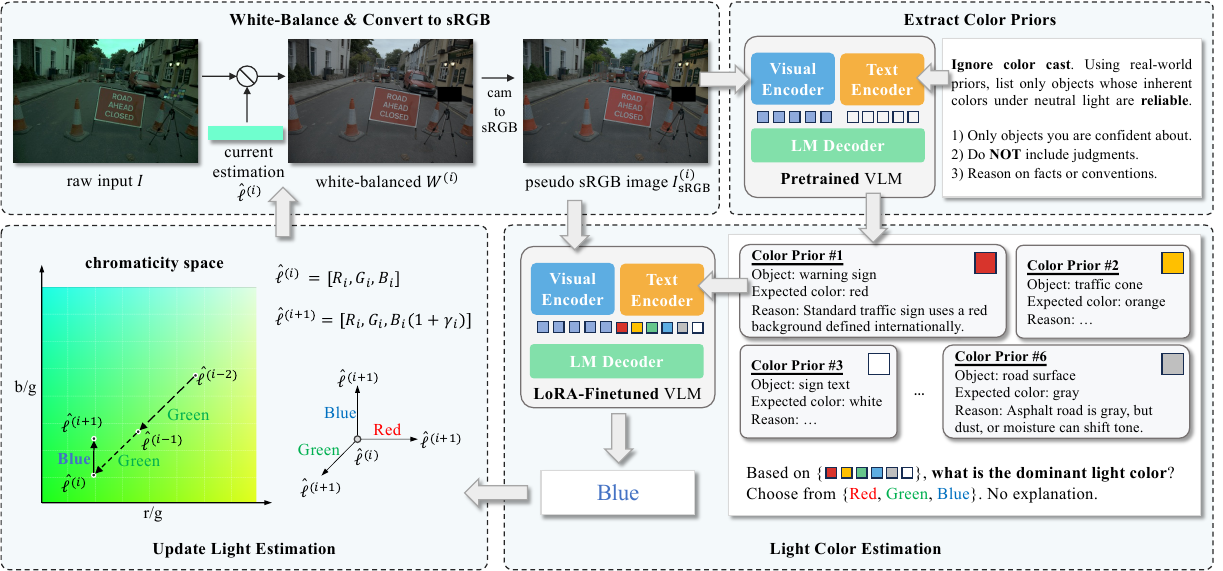}
  \caption{
\textbf{Overview of our proposed framework. }
Given a raw input image, we first apply white balance using the current illuminant estimate and convert the result to sRGB for VLM processing. 
A pretrained VLM extracts semantic color priors from the pseudo-sRGB image, identifying objects whose inherent colors are reliable under neutral light. 
A LoRA-finetuned VLM then predicts the dominant residual light color label (\textit{red}, \textit{green}, or \textit{blue}) based on these priors. 
The illuminant estimate is updated accordingly in chromaticity space and fed back into the next iteration. 
This iterative semantic feedback loop gradually refines the illumination direction until convergence.
}
  \label{fig:overview}
  \vspace{-1em}
\end{figure*}

Learning-based methods~\cite{gehler2008bayesian, gijsenij2010color, bianco2015color, shi2016deep, oh2017approaching, qian2017recurrent, yu2020cascading, barron2015convolutional, barron2017fast, hu2017fc4, xu2020end, lo2021clcc, yue2023color, li2024nightcc, tang2022transfer, afifi2025optimizing,Rizzo_2023,das2021generativemodelsmultiilluminationcolor,10.1007/978-981-95-5702-8_20}, have greatly improved color constancy by predicting illuminants directly from raw images. 
However, their models remain tightly coupled to the training camera’s imaging pipeline, and performance drops when tested on a different sensor. 
Therefore, a new camera often requires additional collected data.

To address this, recent works have explored cross-camera color constancy. 
Some methods utilize limited labeled samples from a new camera, such as Meta-AWB~\cite{mcdonagh2018formulating} and the multi-hypothesis approach~\cite{hernandez2020multi}. DMCC~\cite{yue2024effective} relies on single D65 white-point measurement per sensor. 
Building on convolutional color constancy~\cite{barron2015convolutional}, SIIE~\cite{afifi2019sensor} seeks sensor invariance by mapping raw images to a device-independent working space, while C5~\cite{afifi2021cross} uses unlabeled target-camera RAW in a hypernetwork. 
More recently, generative frameworks such as GCC~\cite{chang2025gcc} synthesize virtual color checkers as an illumination reference. CCMNet~\cite{kim2025ccmnet} leverages pre-calibrated color correction matrices from camera to adapt to the new camera. 
These methods take unwhite-balanced images as input and estimate the illuminant in a single step.
In contrast, we first white-balance the image, then evaluate it in a shared sRGB color space, where we can better exploit the rich semantic information learned during VLM pretraining and iteratively refine the illuminant estimate.

\vspace{0.4em}
\noindent\textbf{Vision-Language Models and Color Understanding }
Recent vision-language models (VLMs)~\cite{ghosh2024exploring,chen2024internvl,bai2025qwen2} integrate a visual encoder with a large language model, enabling open-ended reasoning over images. 
They already show useful color-related abilities: VLMs can associate color terms with visual stimuli, exploit semantic context (object, material, scene), and express color judgments in natural language. 
ColorBench~\cite{liang2025colorbench} reports meaningful performance on several color perception and reasoning tasks, suggesting that non-trivial color priors are encoded in these models. 
Recent studies show that VLMs can organize color space into a shared set of common color terms~\cite{gomez2025color}.

At the same time, prior work suggests that generic VLMs are not yet ideal as plug-and-play tools for precise color estimation. 
They show limited numerical accuracy on color-related queries~\cite{liang2025colorbench}.
Thus, VLMs can provide rich semantic and color priors, but they will benefit from adaptation when used for tasks such as computational color constancy.

\section{Proposed Method}
We assume a single global illuminant, and model each pixel color ${I}\!\in\!\mathbb{R}^{H\times W\times 3}$ as the element-wise product between its white-balanced counterpart ${W}$ and the global illumination color $\boldsymbol{\ell}\!\in\!\mathbb{R}^3$:
\begin{equation}
{I} = {W} \odot \boldsymbol{\ell},
\label{eq:formation}
\end{equation}
where $\odot$ denotes channel-wise multiplication. 
The goal of computational color constancy is to estimate the illuminant $\hat{\boldsymbol{\ell}}$ from the observed RAW image:
$
\hat{\boldsymbol{\ell}} = f({I}).
$

In cross-camera scenarios, raw images are captured by sensors with different spectral sensitivities, leading to distinct distribution of ${I}$. 
Even though Eq.~(\ref{eq:formation}) holds for each sensor, the learned mapping $f$ often fails to generalize when applied to a new camera. 
Our goal is to estimate $\hat{\boldsymbol{\ell}}$ robustly across cameras without per-camera calibration or fine-tuning.

\subsection{Inference Pipeline}
\begin{figure}[t]
    \centering
    \includegraphics[width=\linewidth]{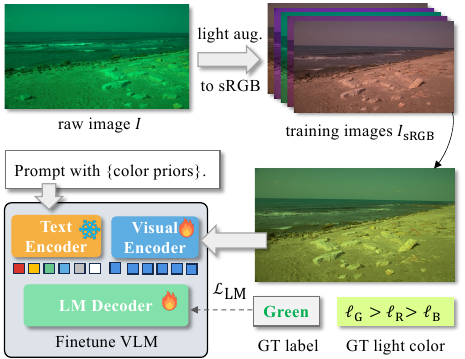}
    \caption{
\textbf{Finetuning pipeline of VLM.} 
Given a raw image, we first apply light-color augmentation in camera color space and convert the results to sRGB. 
These images are processed by a LoRA-finetuned~\cite{lora} VLM, using the same color-prior prompting strategy as in the inference pipeline.
The model predicts the dominant residual light color (\textit{red}, \textit{green}, or \textit{blue}), supervised by the ground-truth illuminant direction. 
A standard language modeling loss $\mathcal{L}_{\mathrm{LM}}$ is applied on the cast label.
}
    \label{fig:train}
\end{figure}
Our inference pipeline follows an iterative perceptual-feedback process.
Each iteration begins by transforming the raw image into a pseudo-sRGB domain, where color judgments align better with the VLM’s pretraining.
Rather than feeding raw sensor data directly to the model, we first apply white balance using the current illuminant estimate $\hat{\boldsymbol{\ell}}^{(t)}$, and then update this estimate based on coarse color-cast feedback predicted by the VLM.

\vspace{0.4em}
\noindent\textbf{White-Balance and sRGB Conversion }
At iteration $t$, given the raw image ${I}$ and the current illuminant estimate $\hat{\boldsymbol{\ell}}^{(t)}\!\in\mathbb{R}^3$, 
the image is first white-balanced as:
\begin{equation}
{W}^{(t)} = {I} \oslash \hat{\boldsymbol{\ell}}^{(t)},
\end{equation}
where $\oslash$ denotes channel-wise division. 
The current white-balanced image ${W}^{(t)}$ is then mapped to an sRGB-like color space by composing a camera-to-XYZ matrix with the standard XYZ-to-sRGB matrix:
\begin{equation}
{I}_{\text{srgb}}^{(t)} = M_{\text{x}\to\text{s}}\, M_{\text{c}\to\text{x}}\, {W}^{(t)}.
\end{equation}
Here $M_{\text{c}\to\text{x}}$ is the $3{\times}3$ camera-to-XYZ matrix (dependent on the sensor’s spectral sensitivities, often referred to as a color correction matrix in camera metadata), and $M_{\text{x}\to\text{s}}$ is the fixed XYZ-to-linear-sRGB matrix.

Note that a standard conversion from camera RGB to sRGB requires accurate white balance, which is not available at this stage.
Our goal is not to obtain a correctly rendered sRGB image, but rather to place the data in a color domain closer to the VLM’s pretraining distribution.
Because VLMs are not trained on large-scale raw sensor data, operating in an sRGB-like space reduces domain shift and stabilizes their color-related predictions.

\vspace{0.4em}
\noindent\textbf{Color Prior Extraction}
Given the initial white-balanced image ${I}_{\text{srgb}}^{(1)}$ and a structured text prompt, 
we feed them to a pretrained vision-language model (VLM) to obtain a formatted color prior list. 
VLM is asked to output a list of 2 to 6 confident objects, each described by {\{object, location, expected color, reason\}}. 

The color prior aims to describe the semantic structure of the scene and provide object-level cues about expected surface colors, which can guide the evaluation across different cameras. 
This prior list will be reused during iterative inference to maintain semantic consistency. 
Because it is derived from the initial white-balanced image ${I}_{\text{srgb}}^{(1)}$, we allow the VLM to perform a \textit{reflection step} after $N$ steps. 
In this step, the VLM re-evaluates the current image ${I}_{\text{srgb}}^{(N)}$ and updates the color prior. 

\vspace{0.4em}
\noindent\textbf{Light Color Estimation }
Given the current sRGB image $I_{\text{srgb}}^{(t)}$ and the color prior, a LoRA-finetuned~\cite{lora} VLM predicts the dominant residual light color of the scene. 
We denote the qualitative prediction as:
\begin{equation}
c^{(t)} = \mathrm{VLM}(I_{\text{srgb}}^{(t)}, \text{prompt}), \quad 
c^{(t)} \in \{\text{red},\,\text{green},\,\text{blue}\},
\end{equation}
where $\text{prompt}$ is the structured prompt inserted with the color prior, including the detected objects, their locations, and expected intrinsic colors. 
The VLM is fine-tuned with LoRA~\cite{lora} adapters applied to the vision tower, language model, and vision-language projector. The training process is detailed in Sec.~\ref{sec:training}.

We classify the residual light color into the three primary directions (red, green, and blue) rather than predicting a continuous RGB value. 
Recent studies \cite{liang2025colorbench} show that VLMs can reliably describe qualitative color attributes but struggle with precise numerical color values. 
This limitation arises because most LLM/VLMs are trained on next-token prediction objectives rather than continuous regression: 
they treat numbers as discrete tokens and are less sensitive to fine-grained numerical variation \cite{li2025exposing,fei2025advancing}. 
Consequently, predicting a categorical color cast aligns better with the model’s linguistic prior and yields more reliable guidance for illuminant estimation across iterations.

\begin{figure*}[t]
  \centering
  \includegraphics[width=\textwidth]{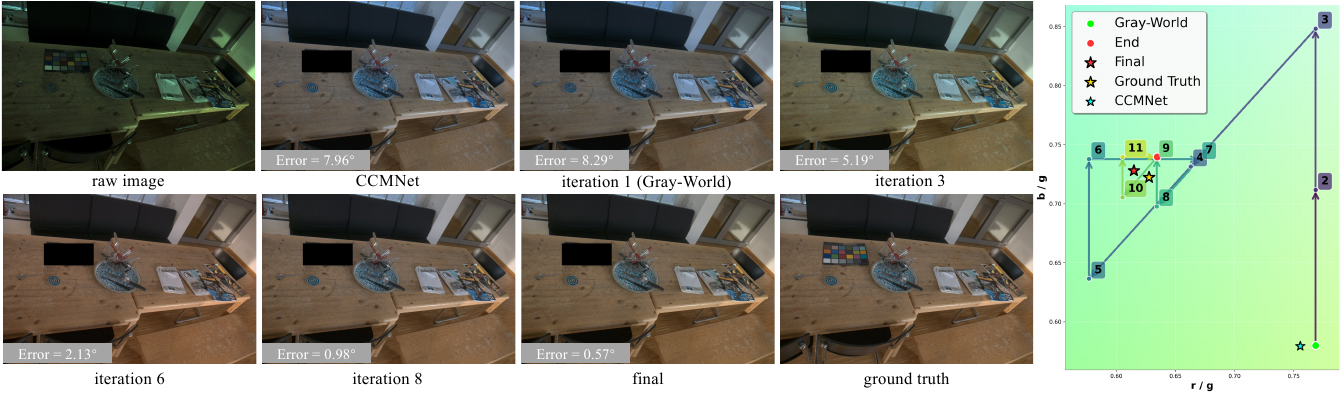}
  \caption{
\textbf{Qualitative example of our iterative correction process. } 
The scene contains large wooden surfaces, which leads CCMNet~\cite{kim2025ccmnet} toward an over-red illuminant estimate. As a result, its white-balanced result appears blue.
Our method starts from Gray-World~\cite{buchsbaum1980spatial} initialization that is also biased by the wood, but iteratively refines the estimate through feedback and converges to $0.57^\circ$. 
The right plot shows the trajectory in chromaticity space. The final illuminant is obtained as the normalized geometric mean of iterations 9, 10, and 11. 
}
  \label{fig:compare}
  \vspace{-1em}
\end{figure*}

\vspace{0.4em}
\noindent\textbf{Update Strategy }
We initialize the illuminant estimate $\hat{\ell}^{(1)}$ using the Gray-World assumption~\cite{buchsbaum1980spatial} for a closer start point. 
The estimate is computed as:
\begin{equation}
\hat{\ell}^{(1)} = 
\mathrm{Normalize}(
\frac{1}{|\Omega|} \sum_{x \in \Omega} I(x)
),
\end{equation}
where $I(x)$ denotes the raw image pixel at location $x$ and $\Omega$ is the image domain. 
This initialization provides a neutral starting point and accelerates convergence during iterative refinement.

For $t>1$, given the current illuminant estimate $\hat{\ell}^{(t)}$ and the predicted color $c^{(t)}$, 
we update the illuminant direction by rotating it toward the corresponding color axis. 
Let $d(c^{(t)}) \in \mathbb{R}^3$ be a unit vector aligned with the predicted cast (red, green, or blue), 
and let
\begin{align}
u^{(t)} &= \mathrm{Normalize}\big(\hat{\ell}^{(t)}\big), \\
v^{(t)} &= \mathrm{Normalize}\big(d(c^{(t)}) -
(d(c^{(t)})^\top u^{(t)})\, u^{(t)}\big).
\end{align}
Given a target step angle $A_t$ (linearly decayed over iterations), 
the updated estimate is
\begin{equation}
\hat{\ell}^{(t+1)} = 
\mathrm{Normalize}\big(\cos A_t \, u^{(t)} + \sin A_t \, v^{(t)}\big).
\label{eq:update}
\end{equation}
This update rotates the illuminant direction by exactly $A_t$ toward $d(c^{(t)})$ in the RGB space.

We monitor the sequence of cast predictions during inference.
As shown in Fig.~\ref{fig:compare}, when three different cast labels appear for the first time, 
we treat this as a coarse convergence signal and reduce all remaining step angles $A_t$ by half, entering a refinement phase.
If three different labels appear again in this refinement phase forming a triangle, or if a fixed iteration limit $T$ is reached, 
the iteration stops.
At termination, the final illuminant is obtained by taking the normalized geometric mean of the last three estimates,
\begin{equation}
\hat{\ell}^{*} =
\mathrm{Normalize}\big(
(\hat{\ell}^{(t)} \odot
 \hat{\ell}^{(t-1)} \odot
 \hat{\ell}^{(t-2)})^{1/3}
\big),
\end{equation}
which stabilizes small oscillations.

\subsection{Training}
\label{sec:training}

We adapt a pretrained vision-language model using LoRA~\cite{lora} to judge the residual cast from an sRGB image and a structured prompt. 
The same prompt format used in inference is adopted here. 
It describes the color prior obtained from a Gray-World-balanced~\cite{buchsbaum1980spatial} image and asks the model to predict the dominant residual cast. 
The model outputs a token from \{red, green, blue\}, conditioned on the image and the prompt. 
Our goal is to link the semantic color priors, which describe object-level intrinsic colors in sRGB space, with the physical illumination direction used to synthesize the training sample.

\paragraph{Data Construction and Augmentation }
For each raw training image with ground-truth illuminant, we first generate a correctly white-balanced version. We then perturb the illuminant direction by a random angle within a range (up to $\sim17.5^\circ$) and reapply it to the image, then convert it to sRGB as described in inference pipeline. This exposes the model to diverse light colors.

\paragraph{LoRA Adaptation and Losses}
We insert LoRA~\cite{lora} adapters into the vision tower, the vision-language projector, and the language model, while keeping the pretrained weights frozen. 
Training uses a standard causal language modeling loss on the answer token.
\begin{equation}
\mathcal{L}_{\text{LM}} = - \sum_{t} \log p_\theta(y_t \mid y_{<t},\, I_{\text{srgb}},\, \text{prompt}),
\end{equation}
where $y_t$ is the $t$-th token of the target sequence, $y_{<t}$ denotes all preceding tokens, and $\theta$ collects the LoRA parameters.
Since the target answer is a single color word (\textit{``red''}, \textit{``green''}, or \textit{``blue''}), the sum reduces to a single cross-entropy term over that token.

\begin{table}
  \centering
  \caption{Leave-one-out evaluation on the Gehler-Shi dataset.}
  \resizebox{\columnwidth}{!}{%
    \begin{tabular}{l|ccccc}
    \toprule
    \multicolumn{6}{c}{{\textbf{Train:} Cube+, NUS-8, Intel-TAU $\rightarrow$ \textbf{Test:} Gehler-Shi}} \\
      \hline
      Method & Mean & Med. & Tri. & B.25\% & W.25\% \\
      \hline
  2nd-order Gray-Edge~\cite{van2007edge}   & 5.13 & 4.44 & 4.62 & 2.11 & 9.26 \\
      Shades-of-Gray~\cite{finlayson2004shades}        & 4.93 & 4.01 & 4.23 & 1.14 & 10.20 \\
      Gray-World~\cite{buchsbaum1980spatial}              & 4.74 & 3.60 & 3.91 & 0.97 & 10.44 \\
      PCA-based B/W Colors~\cite{cheng2014illuminant}  & 3.52 & 2.14 & 2.47 & 0.50 & 8.74 \\
      ASM~\cite{akbarinia2017colour}                   & 3.80 & 2.40 & 2.70 & -    & -    \\
      Woo \emph{et al.}~\cite{woo2017improving}     & 4.30 & 2.86 & 3.31 & 0.71 & 10.14 \\
      Grayness Index~\cite{qian2019cvpr}        & 3.07 & 1.87 & 2.16 & 0.43 & 7.62 \\
      Cross-dataset CC~\cite{koskinen122020cross}      & 2.87 & 2.21 & -    & -    & -    \\
      Quasi-U~\cite{bianco2019quasi}               & 3.46 & 2.23 & -    & -    & -    \\
      SIIE~\cite{afifi2019sensor}                  & 2.77 & 1.93 & -    & 0.55 & 6.53 \\
      FFCC~\cite{barron2017fast}                  & 2.95 & 2.19 & 2.35 & 0.57 & 6.75 \\
      C5 ($m=7$)~\cite{afifi2021cross}            & 2.36 & 1.61 & 1.74 & 0.44 & 5.60 \\
      C5 ($m=9$)~\cite{afifi2021cross}            & 2.50 & 1.99 & 2.03 & 0.53 & 5.46 \\
      CCMNet~\cite{kim2025ccmnet}                & 2.23 & 1.53 & 1.62 & \best{0.36} & 5.46 \\
      \hline
      \textbf{Ours}         & \best{1.52} & \best{1.18} & \best{1.20} & {0.41} & \best{3.29} \\
      \bottomrule
    \end{tabular}
  }
  \label{tab:gehler}
  \vspace{-0.5em}
\end{table}
\begin{table}
  \centering
  \caption{Leave-one-out evaluation on the NUS-8 dataset.}
  \resizebox{\columnwidth}{!}{%
    \begin{tabular}{l|ccccc}
    \toprule
    \multicolumn{6}{c}{{\textbf{Train:} Gehler-Shi, Cube+, Intel-TAU $\rightarrow$ \textbf{Test:} NUS-8}} \\
      \hline
      Method & Mean & Med. & Tri. & B. 25\% & W. 25\% \\
      \hline
      Gray-world~\cite{buchsbaum1980spatial}                 & 4.59 & 3.46 & 3.81 & 1.16 & 9.85 \\
      Shades-of-Gray~\cite{finlayson2004shades}             & 3.67 & 2.94 & 3.03 & 0.98 & 7.75 \\
      Local Surface Refl.~\cite{gao2014efficient}         & 3.45 & 2.51 & 2.70 & 0.98 & 7.32 \\
      PCA-based B/W Colors~\cite{cheng2014illuminant}       & 2.93 & 2.33 & 2.42 & 0.78 & 6.13 \\
      Grayness Index~\cite{qian2019cvpr}             & 2.91 & 1.97 & 2.13 & 0.56 & 6.67 \\
      Cross-dataset CC~\cite{koskinen122020cross}           & 3.08 & 2.24 &   -  &   -  &   -  \\
      Quasi-U~\cite{bianco2019quasi}                     & 3.00 & 2.25 &   -  &   -  &   -  \\
      FFCC~\cite{barron2017fast}                       & 2.87 & 2.14 & 2.30 & 0.71 & 6.23 \\
      C5 ($m=7$)~\cite{afifi2021cross}                 & 2.68 & 2.00 & 2.14 & 0.66 & 5.90 \\
      C5 ($m=9$)~\cite{afifi2021cross}                 & 2.54 & 1.90 & 2.02 & 0.61 & 5.61 \\
      CCMNet~\cite{kim2025ccmnet}                     & 2.32 & 1.71 & 1.83 & 0.53 & 5.18 \\
      \hline
      \textbf{Ours}              & \best{1.83} & \best{1.44} & \best{1.50} & \best{0.51} & \best{3.88} \\
      \bottomrule
    \end{tabular}
  }
  \label{tab:nus_cd}
  \vspace{-0.5em}
\end{table}

\begin{table}
  \centering
  \caption{Leave-one-out evaluation on the Cube+ dataset.}
  \resizebox{\columnwidth}{!}{%
    \begin{tabular}{l|ccccc}
    \toprule
    \multicolumn{6}{c}{{\textbf{Train:} Gehler-Shi, NUS-8, Intel-TAU $\rightarrow$ \textbf{Test:} Cube+}} \\
      \hline
      Method & Mean & Med. & Tri. & B. 25\% & W. 25\% \\
      \hline
      Gray-world~\cite{buchsbaum1980spatial}            & 3.52 & 2.55 & 2.82 & 0.60 & 7.98 \\
      1st-order Gray-Edge~\cite{van2007edge}   & 3.06 & 2.05 & 2.32 & 0.55 & 7.22 \\
      2nd-order Gray-Edge~\cite{van2007edge}   & 3.28 & 2.34 & 2.58 & 0.66 & 7.44 \\
      Shades-of-Gray~\cite{finlayson2004shades}        & 3.22 & 2.12 & 2.44 & 0.43 & 7.77 \\
      Cross-dataset CC~\cite{koskinen122020cross}      & 2.47 & 1.94 & -    & -    & -    \\
      Quasi-U~\cite{bianco2019quasi}                & 2.69 & 1.76 & 2.00 & 0.49 & 6.45 \\
      SIIE~\cite{afifi2019sensor}                  & 2.14 & 1.44 & -    & 0.44 & 5.06 \\
      FFCC~\cite{barron2017fast}                  & 2.69 & 1.89 & 2.08 & 0.46 & 6.31 \\
      DMCC~\cite{yue2024effective}                  & 2.23 & 1.63 & 1.78 & 0.49 & 4.95 \\
      C5 ($m=7$)~\cite{afifi2021cross}            & 1.87 & 1.27 & 1.40 & 0.41 & 4.36 \\
      C5 ($m=9$)~\cite{afifi2021cross}            & 1.92 & 1.32 & 1.46 & 0.44 & 4.44 \\
      CCMNet~\cite{kim2025ccmnet}                & {1.68} & {1.16} & {1.26} & \best{0.38} & {3.89} \\
      \hline
      Ours                  & \best{1.51} & \best{1.09} & \best{1.21} & {0.41} & \best{3.28} \\
      \bottomrule
    \end{tabular}
  }
  \label{tab:cube}
  \vspace{-0.5em}
\end{table}

\begin{table}
  \centering
  \caption{Cross-sensor evaluation on the NUS-8 dataset. Trained on 7 cameras and tested on the last one.}
  \resizebox{\columnwidth}{!}{%
    \begin{tabular}{l|ccccc}
      \toprule
      Method & Mean & Med. & Tri. & B. 25\% & W. 25\% \\
      \hline
      Gray-world~\cite{buchsbaum1980spatial}                 & 4.59 & 3.46 & 3.81 & 1.16 & 9.85 \\
      Shades-of-Gray~\cite{finlayson2004shades}             & 3.67 & 2.94 & 3.03 & 0.98 & 7.75 \\
      Local Surface Refl.~\cite{gao2014efficient}         & 3.45 & 2.51 & 2.70 & 0.98 & 7.32 \\
      Cross-dataset CC~\cite{koskinen122020cross}           & 3.08 & 2.24 &   -  &   -  &   -  \\
      Quasi-U~\cite{bianco2019quasi}                     & 3.00 & 2.25 &   -  &   -  &   -  \\
      PCA-based B/W Colors~\cite{cheng2014illuminant}       & 2.93 & 2.33 & 2.42 & 0.78 & 6.13 \\
      Grayness Index~\cite{qian2019cvpr}             & 2.91 & 1.97 & 2.13 & 0.56 & 6.67 \\
      FFCC~\cite{barron2017fast}                       & 2.87 & 2.14 & 2.30 & 0.71 & 6.23 \\
      DMCC~\cite{yue2024effective}                        & 2.80 & 2.12 & 2.25 & 0.74 & 5.88 \\
      C5~\cite{afifi2021cross}                         & 2.54 & 1.90 & 2.02 & 0.61 & 5.61 \\
      SIIE~\cite{afifi2019sensor}                        & 2.05 & 1.50 &   -  & 0.52 & 4.48 \\
      GCC~\cite{chang2025gcc}                        & 2.03 & 1.78 & 1.83 & 0.77 & 3.69 \\
      C5 (m=9)~\cite{afifi2021cross}                  & 1.77 & 1.37 & 1.46 & 0.48 & 3.75 \\
      CCMNet~\cite{kim2025ccmnet}                     & 1.71 & 1.31 & 1.40 & 0.48 & 3.62 \\
      \hline
      \textbf{Ours}    & \best{1.49} & \best{1.23} & \best{1.28} & \best{0.46} & \best{2.98} \\
      \bottomrule
    \end{tabular}
  }
  \label{tab:nus_cs}
  \vspace{-0.5em}
\end{table}

\begin{table}
  \centering
  \caption{Three-fold cross-validation on the Gehler-Shi dataset. }
  \label{tab:gehler-threefold}

  \resizebox{\columnwidth}{!}{
  \begin{tabular}{l|ccccc}
    \toprule
    {Method} & {Mean} & {Med.} & {Tri.} &
    {B.25\%} & {W.25\%}  \\
    \midrule
    White Patch~\cite{land1977retinex} & 7.55 & 5.68 & 6.35 & 1.45 & 16.12 \\
    Shades-of-Gray~\cite{finlayson2004shades} & 4.93 & 4.01 & 4.23 & 1.14 & 10.20 \\
    Cheng et al.~\cite{cheng2014illuminant} & 3.52 & 2.14 & 2.47 & 0.50 & 8.74 \\
    CNN~\cite{bianco2015color} & 2.36 & 1.95 & -- & -- & -- \\
    MIMT~\cite{li2023mimtmultiilluminantcolorconstancy} & 2.27 & 1.76 & 2.01 & 0.52 & 5.95 \\
    FFCC-2 channels~\cite{barron2017fast} & 1.61 & \best{0.86} & 1.02 & \best{0.23} & 4.27 \\
    SqueezeNet-FC4~\cite{hu2017fc4} & 1.65 & 1.18 & 1.27 & 0.38 & 3.78 \\
    IGTN (full)~\cite{xu2020end} & 1.58 & 0.92 & -- & 0.28 & 3.70 \\
    C4~\cite{hu2017fc4} & 1.35 & 0.88 & 0.99 & 0.28 & 3.21 \\
    CLCC~\cite{lo2021clcc} & 1.44 & 0.92 & 1.04 & 0.27 & 3.48 \\ 
    GCC~\cite{chang2025gcc} & 1.91 & 1.80 & 1.84 & 0.60 & 3.46 \\ \hline
    Ours & \best{1.34} & {0.92} & \best{0.98} & 0.29 & \best{3.13} \\
    \bottomrule
  \end{tabular}}
  \label{tab:supp-gehler}
  \vspace{-1em}
\end{table}

\section{Experiments}
\subsection{Implementation Details}
We use Qwen2.5-VL 7B~\cite{bai2025qwen2} as the pretrained VLM backbone, 
due to its strong visual recognition and color-reasoning capabilities~\cite{liang2025colorbench}.
All base weights remain frozen; only LoRA~\cite{lora} adapters are trained.
LoRA~\cite{lora} modules are inserted into all layers of the vision tower, 
the vision-language projector, and the language model, 
with rank $r=8$.
As the full prompts are long, we include the complete format in the supplementary material.

For each raw training image, we apply our light-color augmentation, 
convert the result to sRGB, and resize it so that the shorter side is 448 pixels.
We train only the LoRA~\cite{lora} parameters for 800 iterations using AdamW~\cite{adamw}, 
with an effective batch size of 512 (physical batch size 16 and gradient accumulation of 32) 
and a learning rate of $4\times10^{-4}$.
All training is conducted on a single NVIDIA H200 GPU.
At test time, we run at most $T=20$ iterations.
The step angle decays linearly from $3^\circ$ to $0.1^\circ$.

\begin{table*}[t]
  \centering
  \caption{Camera-agnostic evaluation. All results are in degrees.}
  \resizebox{\textwidth}{!}{
    \begin{tabular}{l|ccccc|ccccc}
      \toprule
      \multirow{2}{*}{Method} & \multicolumn{5}{c|}{\textbf{Train:} NUS-8$\rightarrow$\textbf{Test:} Gehler-Shi} & \multicolumn{5}{c}{\textbf{Train:} Gehler-Shi$\rightarrow$\textbf{Test:} NUS-8} \\
       & Mean & Median & Tri-mean & Best 25\% & Worst 25\% & Mean & Median & Tri-mean & Best 25\% & Worst 25\% \\
      \hline
      \multicolumn{11}{l}{\textbf{Statistical Methods}} \\
      \hline
      White-Patch~\cite{land1977retinex}              & 7.55 & 5.68 & 6.35 & 1.45 & 16.12 & 9.91 & 7.44 & 8.78 & 1.44 & 21.27 \\
      Gray-World~\cite{buchsbaum1980spatial}               & 6.36 & 6.28 & 6.28 & 2.33 & 10.58 & 4.59 & 3.46 & 3.81 & 1.16 & 9.85 \\
      1st-order Gray-Edge~\cite{van2007edge}      & 5.33 & 4.52 & 4.73 & 1.86 & 10.43 & 3.35 & 2.58 & 2.76 & 0.79 & 7.18 \\
      2nd-order Gray-Edge~\cite{van2007edge}      & 5.13 & 4.44 & 4.62 & 2.11 & 9.26  & 3.36 & 2.70 & 2.80 & 0.89 & 7.14 \\
      Shades-of-Gray~\cite{finlayson2004shades}           & 4.93 & 4.01 & 4.23 & 1.14 & 10.20 & 3.67 & 2.94 & 3.03 & 0.99 & 7.75 \\
      General Gray-World~\cite{barnard2002comparison}       & 4.66 & 3.48 & 3.81 & 1.00 & 10.09 & 3.20 & 2.56 & 2.68 & 0.85 & 6.68 \\
      Gray Pixel (edge)~\cite{qian2018revisiting}        & 4.60 & 3.10 &  -   &  -   &  -    & 3.15 & 2.20 &  -   &  -   &  -    \\
      Cheng et al.~\cite{cheng2014illuminant}             & 3.52 & 2.14 & 2.47 & 0.50 & 8.74  & 2.92 & 2.04 & 2.24 & 0.62 & 6.61 \\
      LSRS~\cite{gao2014efficient}                     & 3.31 & 2.80 & 2.87 & 1.14 & 6.39  & 3.45 & 2.51 & 2.70 & 0.98 & 7.32 \\
      GI~\cite{qian2019cvpr}                       & 3.07 & 1.87 & 2.16 & \best{0.43} & 7.62  & 2.91 & 1.97 & 2.13 & \best{0.56} & 6.67 \\
      \hline
      \multicolumn{11}{l}{\textbf{Learning-based Methods}} \\
      \hline
      Bayesian~\cite{gehler2008bayesian}                 & 4.75 & 3.11 & 3.50 & 1.04 & 11.28 & 3.65 & 3.08 & 3.16 & 1.03 & 7.33 \\
      Chakrabarti~\cite{chakrabarti2015color}              & 3.52 & 2.71 & 2.80 & 0.86 & 7.72  & 3.89 & 3.10 & 3.26 & 1.17 & 7.95 \\
      FFCC~\cite{barron2017fast}                     & 3.91 & 3.15 & 3.34 & 1.22 & 7.94  & 3.19 & 2.33 & 2.52 & 0.84 & 7.01 \\
      SqueezeNet-FC4~\cite{hu2017fc4}           & 3.02 & 2.36 & 2.50 & 0.81 & 6.36  & 2.40 & 2.03 & 2.10 & 0.70 & 4.80 \\
      C$^4$-SqueezeNet-FC4~\cite{hu2017fc4}     & 2.73 & 2.20 & 2.28 & 0.72 & 5.69  & 2.28 & 1.90 & 1.97 & 0.67 & 4.60 \\
      SIIE~\cite{afifi2019sensor}                     & 3.72 & 2.46 & 2.79 & 1.02 & 8.51  & 4.24 & 3.88 & 3.93 & 1.45 & 7.66 \\
      CLCC~\cite{lo2021clcc}                     & 3.62 & 2.57 & 2.51 & 0.89 & 6.30  & 3.42 & 2.95 & 3.06 & 0.94 & 6.70 \\
      C5~\cite{afifi2021cross}                    & 3.34 & 2.57 & 2.68 & 0.78 & 7.39  & 2.65 & 1.98 & 2.14 & 0.66 & 5.72 \\
      GCC~\cite{chang2025gcc}                      & 2.35 & 2.02 & 2.06 & 0.78 & 4.57  & 2.38 & 2.01 & 2.10 & 0.80 & 4.58 \\
      CCMNet~\cite{kim2025ccmnet} & 2.38 & \best{1.74} & 1.88 & 0.46 & 5.62 &2.17 & \best{1.65} & \best{1.76} & 0.59 & 4.75 \\
      \hline
      Ours & \best{2.03} & \best{1.74} & \best{1.79} & 0.67 & \best{4.06} & \best{2.07} & {1.72} & \best{1.76} & 0.62 & \best{4.11} \\
      \bottomrule
    \end{tabular}
  }
  \vspace{-1em}
  \label{tab:nus-gehler}
\end{table*}

\subsection{Datasets and Protocol}

We follow the cross-camera evaluation setup commonly used in recent work~\cite{afifi2021cross,kim2025ccmnet,afifi2019sensor}. 
We use four public RAW datasets: Gehler-Shi~\cite{gehler2008bayesian,hemrit2018rehabilitating} 
(568 images from Canon 1D and 5D), 
NUS-8~\cite{cheng2014illuminant} (1{,}736 images from eight cameras, each with a color chart for GT), 
Intel-TAU~\cite{inteltau} (7{,}022 images from Canon 5DSR, Nikon D810, and Sony IMX135), 
and Cube+~\cite{ershov2020cube++} (2{,}070 images from a Canon 550D).

Unless stated otherwise, we adopt a leave-one-out cross-dataset protocol: for each target dataset, the model is trained on the remaining datasets and evaluated on the held-out dataset, ensuring no camera overlap. Following CCMNet~\cite{kim2025ccmnet}, we exclude the Sony IMX135 subset of Intel-TAU since its color correction matrix (CCM) is unavailable, which prevents camera-to-sRGB conversion. The remaining Intel-TAU images are used only for training, and other CCMs are obtained via Adobe DNG Converter.
We additionally report the standard cross-sensor (CS) protocol on NUS-8, where one camera is used for testing and the remaining seven for training, as well as cross-dataset generalization between NUS-8 and Gehler-Shi. Moreover, we evaluate VLM-CC using the standard threefold cross-validation protocol on Gehler-Shi dataset.

We report standard angular-error statistics between estimated and ground-truth illuminants: 
mean, median, trimean, mean of Best-25\%, and mean of Worst-25\%.

\subsection{Results}
Our method consistently achieves the best performance across all held-out evaluation settings.
Across Gehler-Shi (Table~\ref{tab:gehler}), NUS-8 cross-dataset (Table~\ref{tab:nus_cd}), and Cube+ (Table~\ref{tab:cube}), nearly all evaluation metrics are improved compared with prior learning-based methods. 
Notably, the largest improvements are observed on the Worst-25\% metric, indicating that our approach substantially reduces large-error cases and improves robustness in challenging scenes.
Under the NUS-8 cross-sensor protocol (Table~\ref{tab:nus_cs}), a similar trend is observed, with improvements across all reported metrics.

In another cross-camera setting (NUS-8 $\rightarrow$ Gehler and Gehler $\rightarrow$ NUS-8; Table~\ref{tab:nus-gehler}), our model again achieves the lowest errors in nearly all metrics for both transfer directions among learning-based methods. 
Unlike the other cross-dataset experiments, this setting uses only 1{,}736 images from NUS-8 and 568 images from Gehler-Shi. The results suggest that, with the help of semantic priors, our method remains effective even under limited data.

Although cross-camera and cross-dataset generalization are our main focus, we also evaluate VLM-CC under the standard threefold cross-validation protocol on Gehler-Shi in Table~\ref{tab:gehler-threefold}. The results remain competitive with strong single-illuminant baselines, while also surpassing the recent cross-camera method GCC~\cite{chang2025gcc}.

\paragraph{Discussion }
Across nearly all evaluations, our method consistently outperforms prior work.
This holds across diverse settings, from cross-sensor to fully cross-camera evaluation and from multi-dataset training to the single-dataset regime. 
Our semantic-driven feedback-based framework consistently reduces Worst-25\% errors, highlighting its robustness under challenging illuminant conditions. Moreover, this advantage becomes even more obvious as the training data becomes more diverse. For example, on Gehler-Shi, when moving from training only on NUS-8 (Table~\ref{tab:nus-gehler}) to  Cube+ + NUS-8 + Intel-TAU (Table~\ref{tab:gehler}), a strong baseline, CCMNet’s mean error improves from 2.38° to 2.23° (about an 9\% reduction), whereas ours improves from 2.03° to 1.52° (about a 26\% reduction), with Worst-25\% dropping from 3.91° to 3.08°. These observations suggest that our framework scales more effectively with increasing data diversity, indicating a greater performance ceiling when sufficient training data is available.

\subsection{Ablation Study}
\noindent\textbf{Direct RGB Illuminant Regression }
Table~\ref{tab:ablation-all}~(a) compares three different color constancy strategies using VLM. The one-step baseline directly predicts the RGB illuminant from the
sRGB input in a single forward pass. The performance shows that
cross-camera illuminant regression is challenging for a VLM even with LoRA
adaptation. The second variant applies our iterative pipeline but asks the VLM
to output numerical RGB values after each white-balance step. This strategy significantly improves
accuracy, showing that ``white-balance first, adjust later'' is a stronger
formulation. Our full method further improves all metrics by asking the VLM to predict only a discrete cast label at each iteration, which better matches its strength in categorical color judgments and avoids unstable numerical regression.

\begin{table}
  \centering
  \caption{Comprehensive ablation study in leave-one-out evaluation on Gehler-Shi dataset.}
  \renewcommand{\arraystretch}{1.1}
  \resizebox{\columnwidth}{!}{%
    \begin{tabular}{l|ccccc}
      \toprule
      \multicolumn{6}{c}{\textbf{Train:} Cube+, NUS-8, Intel-TAU $\rightarrow$ \textbf{Test:} Gehler-Shi} \\
      \hline
      Method & Mean & Med. & Tri. & B.25\% & W.25\% \\
      \hline
      
      \multicolumn{6}{c}{\textbf{(a) Inference Strategy}} \\[-1pt]
      \hline
      one-step numerical        & 3.59 & 2.72 & 2.90 & 1.12 & 7.22 \\
      iterative numerical       & 1.70 & 1.36 & 1.37 & 0.73 & 3.68 \\
      iterative discrete (ours)                       & \textbf{1.52} & \textbf{1.18} & \textbf{1.20} & \textbf{0.41} & \textbf{3.29} \\
      \hline
      
      \multicolumn{6}{c}{\textbf{(b) Initialization Method}} \\[-1pt]
      \hline
      w/o initialization  & 1.61 & 1.27 & 1.29 & 0.47 & 3.60 \\
      2nd-order Gray-Edge & 1.58 & 1.32 & 1.39 & 0.50 & \textbf{3.29} \\
      Gray-World (ours)          & \textbf{1.52} & \textbf{1.18} & \textbf{1.20} & \textbf{0.41} & \textbf{3.29} \\
      \hline
      
      \multicolumn{6}{c}{\textbf{(c) VLM Architecture and Size}} \\[-1pt]
      \hline
      InternVL-3.5 1B            & 1.73 & 1.20 & 1.25 & 0.44 & 3.87 \\
      Qwen2.5-VL 3B              & 1.71 & 1.44 & 1.49 & 0.66 & 3.47 \\
      Qwen2.5-VL 7B (ours)              & \textbf{1.52} & \textbf{1.18} & \textbf{1.20} & {0.41} & \textbf{3.29} \\
      \hline
      
      \multicolumn{6}{c}{\textbf{(d) Semantic and Spatial Cues}} \\[-1pt]
      \hline
      random color priors       & 1.93 & 1.62 & 1.57 & 0.56 & 4.88 \\
      shuffled input images             & 2.16 & 1.55 & 1.57 & 0.47 & 5.27 \\
      ours                       & \textbf{1.52} & \textbf{1.18} & \textbf{1.20} & \textbf{0.41} & \textbf{3.29} \\
      \hline
      
      \multicolumn{6}{c}{\textbf{(e) Finetuned Modules}} \\[-1pt]
      \hline
      w/o finetuning      & 14.33 & 13.21 & 13.18 & 7.19 & 23.29 \\
      LM decoder                 & 2.12  & 1.71  & 1.83  & 0.62 & 5.41 \\
      vision tower        & 1.77  & 1.44  & 1.39  & 0.53 & 4.01 \\
      LM \& vision tower (ours)  & \textbf{1.52} & \textbf{1.18} & \textbf{1.20} & \textbf{0.41} & \textbf{3.29} \\
      \bottomrule
    \end{tabular}
  }
  \label{tab:ablation-all}
  \vspace{-1em}
\end{table}

\vspace{0.4em}
\noindent\textbf{Robustness to Initial Estimation }
We compare different initialization strategies in Table~\ref{tab:ablation-all}~(b). 
Replacing the initial illuminant from Gray-World~\cite{buchsbaum1980spatial} to 2nd-order Gray-Edge~\cite{van2007edge}, or even removing the initialization step, leads to small change in performance. 
This suggests that the iterative feedback process effectively corrects the initial bias and converges to a stable solution. 
The framework therefore relies little on the accuracy of the starting point, showing strong robustness to imperfect initialization.

\vspace{0.4em}
\noindent\textbf{VLM Model and Size }
Table~\ref{tab:ablation-all}~(c) evaluates our framework with different VLM backbones and model sizes. The results show that the approach works consistently across different architectures~\cite{bai2025qwen2, chen2024internvl} and parameter numbers. We observe that different VLM models may exhibit differences in performance, reflecting variations in their underlying capabilities. In addition, within the same model family, larger variants lead to slightly better results. That said, while the choice and scale of the VLM can influence performance to some extent, the proposed framework remains robust across different models and sizes without architectural changes.

\vspace{0.4em}
\noindent\textbf{Semantic and Spatial Priors}
As shown in Table~\ref{tab:ablation-all}~(d), semantic cues are essential for stable
illuminant estimation. Using random priors, where object names and expected colors are randomly generated, degrades performance. This shows that arbitrary color priors cannot guide the VLM to a correct color direction. When we shuffle the image by splitting it into $14\times14$ patches and randomly permuting them, the model receives a severely distorted structure and usually cannot generate color priors on it. Interestingly, the Best-25\% error changes little, whereas the Worst-25\% error
increases notably. We hypothesize that for images that are already easy to correct (small angular
errors), the model can still rely on statistic information. For worst-25\% cases, semantic cues and spatial consistency become more important in the inference.

\vspace{0.4em}
\noindent\textbf{Effect of Finetuning }
Table~\ref{tab:ablation-all}~(e) shows that finetuning is essential for our VLM-based
illuminant estimation. Without any adaptation, the pretrained VLM performs poorly because
neither the vision tower nor the language model has been specifically trained to interpret color casts.
Finetuning only the language model significantly improves accuracy by aligning the extracted
visual features and color priors with the discrete illuminant labels.
Finetuning the vision tower further improves performance, as the pretrained visual
encoder is not optimized for fine-grained chromatic cues over significant unwhite-balanced images. Finetuning both components yields the best results.

\section{Conclusion}
We introduced \textbf{VLM-CC}, a feedback-driven framework for cross-camera color constancy that leverages a vision-language model to guide iterative illuminant refinement. 
Rather than estimating illumination directly from raw inputs, VLM-CC first performs white balance, maps the result into a pseudo-sRGB domain, and uses VLM-derived feedback to update the estimate over successive iterations. 
Across four benchmark datasets, our method achieves state-of-the-art cross-camera performance and substantial reductions in worst-case errors. 
Ablation studies further show that semantic cues from the VLM, iterative correction, and targeted LoRA finetuning each contribute meaningfully to the observed robustness. 
Overall, VLM-CC demonstrates that VLM-based perceptual feedback can serve as an effective mechanism for improving sensor generalization in computational color constancy.

\section*{Acknowledgments}
This document is supported by the Ministry of Education, Singapore, under its MOE AcRF TIER 3 Grant (MOE-MOET32022-0001).

{
    \small
    \bibliographystyle{ieeenat_fullname}
    \bibliography{combined_fixed}
}

\end{document}

%% file: preamble.tex
\PassOptionsToPackage{table}{xcolor}
\usepackage{algpseudocode}
\usepackage{graphicx}

\usepackage{amsmath}
\usepackage{amssymb}
\usepackage{booktabs}
\usepackage{bbding}
\usepackage{soul}
\usepackage{bbm}
\usepackage{multirow}
\usepackage[accsupp]{axessibility}
\usepackage{xcolor}
\usepackage{colortbl}
\usepackage{makecell}

\usepackage{graphicx}


\usepackage[ruled,vlined]{algorithm2e}
\usepackage{amsmath}

\usepackage{booktabs}
\usepackage{subcaption}

\definecolor{bestbg}{RGB}{232,246,239} 
\newcommand{\best}[1]{\cellcolor{bestbg}\textbf{#1}}

\usepackage{listings}
\usepackage{xcolor}

\usepackage{multirow}